# Mutual Information Tracks Policy Coherence in Reinforcement Learning

Cameron Reid
cameron.reid@semarx.com
Semarx Research Ltd.
Alexandria, VA, USA

Wael Hafez
wael.hafez@semarx.com
Semarx Research Ltd.
Alexandria, VA, USA

Amir Nazeri
amir.nazeri@semarx.com
Semarx Research Ltd.
Alexandria, VA, USA

*Abstract*—**Reinforcement Learning (RL) agents deployed in real-world environments face degradation from sensor faults, actuator wear, and environmental shifts, yet lack intrinsic mechanisms to detect and diagnose these failures. We present an information-theoretic framework that reveals both the fundamental dynamics of reinforcement learning and provides practical methods for diagnosing deployment-time anomalies. Through analysis of state-action mutual information patterns in a robotic control task, we first demonstrate that successful learning exhibits characteristic information signatures: mutual information between states and actions steadily increases from 0.84 to 2.83 bits (238% growth) despite growing state entropy, indicating that agents develop increasingly selective attention to task-relevant patterns. Intriguingly, states, actions and next states joint mutual information, MI(S,A;S'), follows an inverted U-curve, peaking during early learning before declining as the agent specializes—suggesting a transition from broad exploration to efficient exploitation. More immediately actionable, we show that information metrics can differentially diagnose system failures: observation-space, i.e., states noise (sensor faults) produces broad collapses across all information channels with pronounced drops in state-action coupling, while action-space noise (actuator faults) selectively disrupts action-outcome predictability while preserving state-action relationships. This differential diagnostic capability—demonstrated through controlled perturbation experiments—enables precise fault localization without architectural modifications or performance degradation. By establishing information patterns as both signatures of learning and diagnostic for system health, this work provides the foundation for adaptive RL systems capable of autonomous fault detection and policy adjustment based on information-theoretic principles.**

## I. INTRODUCTION

Reinforcement learning (RL) has advanced dramatically, enabling systems to master complex control tasks from robotics to game playing [27]. Despite these successes, our understanding of how agents develop and maintain effective representations under varying environmental complexity remains limited. Current performance metrics like reward accumulation or value loss provide little insight into whether an agent has developed robust representations or merely memorized specific state-action mappings [41]. This gap becomes critical when agents encounter environmental shifts, component degradation, or novel scenarios, often resulting in unexpected performance collapse without warning [43]. Traditional approaches to analyzing representation quality rely on model-specific methods such as attention visualization [36] or feature importance scoring [18]. However, these approaches generally require special architectural modifications, struggle to provide quantitative metrics of representation quality, and offer little predictive power regarding when policies might degrade. As RL systems move into safety-critical applications, developing universal, interpretable measures of representation quality becomes increasingly vital [2].

### A. Information-Theoretic Metrics as Tools for Policy Analysis

Information theory offers a principled framework for analyzing representation quality independent of specific architectural choices [7]. Several recent works have applied information-theoretic concepts to reinforcement learning, primarily focusing on training dynamics. Minimum entropy policies encourage exploration through maximizing action entropy [16], while information bottleneck approaches aim to discard irrelevant state information [14]. However, these approaches generally treat information measures as optimization targets rather than analytical tools. More recently, Anand et al. [3], applied mutual information to analyze representation compression in deep RL networks, while Igl et al. [19], used information-theoretic regularization to improve generalization. Despite these advances, information theory remains underutilized for monitoring and analyzing policy quality post-training. In particular, tracking mutual information between states and actions during deployment offers a potentially powerful, architecture-agnostic approach to measuring how effectively an agent's policy captures task-relevant patterns [5]– an approach we develop in this work. Moreover, these same information patterns that reveal learning dynamics may also serve as sensitive indicators of system degradation, potentially enabling differential diagnosis of failure modes based on how information flow disrupts across different channels.

## B. Our contribution

This paper introduces mutual information between states and actions as a quantitative metric for assessing representation quality in reinforcement learning agents. Through systematic analysis of a robotic arm control task, we demonstrate that successful learning manifests as increasing mutual information between states and actions despite growing state entropy – revealing that effective agents develop increasingly selective attention to task-relevant patterns. Specifically, we show that MI(S;A) increases from 29.8% to 37.3% relative to state entropy even as the agent encounters more diverse states, indicating strengthening state-action relationships rather than mere memorization. This metric proves valuable for detecting misalignment between policies and environments before performance visibly degrades, providing early warning of potential failures. Unlike specialized architectural approaches to representation learning, our information-theoretic analysis applies to standard RL algorithms without modification, offering a universal lens for understanding representation development in diverse learning systems. These findings establish information throughput as both a lens for understanding how representations develop during learning and a diagnostic tool for maintaining system reliability during deployment, demonstrating that the same information patterns that reveal learning dynamics can detect and localize system failures [32] [21].

## II. RELATED WORK

### A. Representation Learning in Reinforcement Learning

Developing effective state representations is crucial for reinforcement learning performance and generalization. Early approaches focused on hand-crafted features [Sutton & Barto, 1998], while deep RL enabled learning representations directly from raw observations [27]. Recent advances have focused on auxiliary objectives to improve representation quality without changing the core RL algorithm. François-Lavet et al. [9] demonstrated that predicting future states improves sample efficiency, while Jaderberg et al. [20], showed that multiple auxiliary tasks enhance representation learning in the Atari domain. Contrastive learning has emerged as a powerful approach for representation quality, with Laskin et al. [24] introducing CURL, which applies contrastive methods to RL, significantly improving sample efficiency. Similarly, Stooke et al. [33] proposed decoupling representation learning from policy optimization, enabling more stable learning in complex environments. Despite these advances, most methods focus on improving representations during training rather than analyzing their quality during deployment. As Wang et al. [17], note, representation collapse remains a significant challenge, where agents develop brittle features that fail to capture task-relevant information outside specific training conditions, highlighting the need for quantitative metrics of representation quality.

### B. Information Theory in Policy Evaluation and Learning

Information theory has provided valuable tools for understanding and improving reinforcement learning algorithms. Tishby et al. 0 introduced the Information Bottleneck principle, later extended to deep learning by Tishby & Zaslavsky [37], proposing that neural networks optimize the tradeoff between compression and prediction. In RL specifically, Galashov et al. [11], applied information-theoretic constraints to improve transfer learning, while Goyal et al. [14], used an information bottleneck approach to enhance exploration. Mutual information estimation in high-dimensional spaces presents unique challenges addressed by Belghazi et al. [5], who introduced neural estimators that enable practical application to complex systems. Strouse et al. [34], explored how mutual information between states and actions reflects intrinsic motivation in exploration, while Abel et al. [1], used information-theoretic measures to quantify task complexity in RL environments. More directly related to our work, Lu et al. [26] explored how predictive information structures evolve during RL training to improve sample efficiency. However, existing literature primarily treats information measures as training objectives or analysis tools during development rather than as monitoring metrics during deployment, creating an opportunity for real-time assessment of policy quality using information-theoretic principles."

### C. Detecting Distribution Shifts in Deployed RL Systems

Detecting when reinforcement learning systems operate outside their training distribution remains an open challenge with significant safety implications. Amodei et al. [2], highlighted this as a key concern for real-world RL deployment, while Dulac-Arnold et al. [8], identified reliable operation under distribution shift as one of nine challenges for real-world reinforcement learning. Traditional approaches for detecting distribution shifts include uncertainty estimation through ensemble methods [23] and Bayesian neural networks [10], though these typically require architectural modifications. Rabanser et al. [29], proposed statistical tests for detecting dataset shifts but focused primarily on supervised learning settings.

In the RL domain, Zhang et al. [41], developed methods for detecting out-of-distribution states based on prediction errors from learned dynamics models, while Sedlmeier et al. [30], explored uncertainty-aware policies that explicitly model distributional shifts during execution. Most approaches, however, rely on comparing current observations to training data or require specialized architectures. Kumar et al. [22] noted that most techniques fail to detect subtle distribution shifts until performance has already degraded significantly, highlighting the need for more sensitive metrics capable of identifying misalignment between learned policies and current environments before failures occur.

Beyond merely detecting that something has gone wrong, a critical challenge for deployed RL systems is diagnosing the specific nature of the failure. Current approaches typically treat all distribution shifts as monolithic events, unable to distinguish whether degradation stems from sensor faults, actuator wear, or environmental changes. This diagnostic limitation prevents targeted interventions - a system that could differentiate between perception failures and action execution failures could potentially adapt its behavior accordingly. To our knowledge, no existing method provides such differential diagnosis of failure modes in RL systems, representing a significant gap in our ability to build truly resilient autonomous agents.

## III. METHODOLOGY

### A. Overview of Experimental Design

We conducted a two-phase experimental study to evaluate how information-theoretic metrics can both reveal learning dynamics and diagnose system failures in reinforcement learning agents. In the first phase, we tracked entropy and mutual information metrics throughout 200,000 training steps as an agent learned a robotic manipulation task. This allowed us to identify characteristic information patterns that emerge during different stages of learning.

In the second phase, we tested the diagnostic capabilities of these same metrics by introducing controlled perturbations to the fully-trained agent. We injected noise into either the observation stream (simulating sensor faults) or action outputs (simulating actuator degradation) and monitored how information flow patterns changed in response. By using the same agent and task across both phases, we could directly connect learning dynamics to vulnerability patterns, demonstrating that information metrics provide a unified framework for both understanding and monitoring RL systems.

### B. Environment and Task Description

We used the Reach task from the panda-gym environment [12], where a Panda robotic arm is controlled through inverse kinematics with the RL agent providing end-effector displacement commands in Cartesian coordinates ($\Delta x$, $\Delta y$, $\Delta z$). The workspace spans [-0.5, 0.5] meters in each dimension, with target positions randomly sampled within this space. Episodes terminate successfully when the end-effector reaches within 0.1 units of the target or after 500 timesteps.

To enable information-theoretic analysis, we discretized the continuous spaces as follows: the 3D distance-to-target vector was discretized into 10 bins per dimension (yielding 1,000 possible states), while action commands were discretized into 7 levels per dimension (343 possible actions). This discretization resolution was chosen to balance information sensitivity with statistical reliability given our dataset size.

### C. Learning Algorithm and Training Protocol

We trained the agent using Proximal Policy Optimization (PPO) [30], implemented in stable-baselines3. PPO was chosen for its stability and sample efficiency in continuous control tasks. The key hyperparameters are: clip range $\varepsilon$=0.2, discount factor $\gamma$=0.99, GAE $\lambda$=0.95, learning rate $3\times10^{-4}$, with 10 epochs per update and batch size of 64. The policy network consisted of two hidden layers with 64 units each using ReLU activations.

Training ran for 200,000 environment steps. Every timestep we logged the full (state, action, next-state) tuple, producing $\approx$ 200 000 transitions for information analysis. We executed four random-seed replicates and confirmed qualitatively identical information signatures; to avoid redundancy we report the detailed results from one representative seed, which suffice for the cumulative-window characterization.

### D. Information-Theoretic Metrics and Computation

We calculated three primary information-theoretic metrics to analyze agent behavior. First, we computed Shannon entropy $H(X) = -\Sigma\ p(x)\ \log_2\ p(x)$ for states, actions, and next states, where probabilities were estimated by counting occurrences in our discretized spaces. Second, we calculated mutual information $MI(S;A) = H(S) + H(A) - H(S,A)$ to measure how strongly states and actions are coupled. Third, we computed the joint mutual information $MI(S,A;S') = H(S,A) + H(S') - H(S,A,S')$, which quantifies the total information the current state-action pair provides about the next state. For the perturbation analysis, we additionally calculated $MI(A;S')$ and $MI(S;S')$ to provide more granular insights into which specific information channels were disrupted by sensor versus actuator noise.

Specifically, for training analysis we used cumulative windows of 5,000 environment steps, with counts accumulated across all previous windows to capture the convergence of the joint state-action distribution as the policy stabilized. For deployment monitoring, and perturbations analysis we used 2,000-step sliding windows.

### E. Perturbation Experiments

To evaluate the diagnostic capabilities of our information metrics, we conducted controlled perturbation experiments on the fully-trained agent. After completing 200,000 training steps, we froze the policy (disabled gradient updates) and deployed it for 20,000 additional steps. At step 10,000, we introduced continuous perturbations in separate experimental runs: (1) observation noise, where Gaussian noise ($\sigma^2 = 0.1$) was continuously added to state observations to simulate persistent sensor degradation, or (2) action noise, where Gaussian noise ($\sigma^2 = 0.1$ and $\sigma^2 = 1.0$) was continuously added to action outputs to simulate ongoing actuator faults.

We computed information metrics using 2,000-step sliding windows to compare system behavior before and after perturbation onset. The pre-perturbation baseline was established from steps 8,000-10,000, while post-perturbation metrics were tracked from step 10,000 onward. This approach allowed us to identify rapid changes in information flow patterns characteristic of each failure type. The continuous noise injection simulates realistic failure modes where sensors or actuators degrade persistently rather than experiencing isolated glitches.

### F. Windowing Strategy: Learning vs. Monitoring

During training we treat the run as a single, growing system: each new state–action–next-state tuple permanently expands the agent's experience and the policy it is learning. To capture how the total information structure of that system matures, we compute entropies and mutual-information metrics cumulatively—each window adds to all previous data. The resulting curves show how the policy's overall coherence increases (or stalls) as learning proceeds.

In deployment/monitoring, the goal flips from measuring growth to detecting change. Here we are no longer interested in the lifetime statistics of the policy but in whether its current behavior has drifted from its baseline. We therefore use a fixed-size sliding window (2,000 steps): each metric is calculated only on the most recent window, which "slides" forward step-by-step. This makes the information signature sensitive to abrupt shifts—such as sensor faults or injected noise—while ignoring distant, now-irrelevant history. In short, cumulative windows reveal learning progress; sliding windows reveal operational deviations.

## IV. RESULTS

### A. Information Dynamics During Training

We evaluated our information-theoretic approach using a reaching task where an agent learns to reach efficiently through a discretized 3D space. The state space consists of the agent's current cell position in the 3D grid, while actions represent movement directions between adjacent cells. The agent's goal was to navigate from its current cell to a specified target cell using the most efficient path.

We tracked information metrics throughout 200,000 training steps, calculating entropy and mutual information directly from the frequencies of state-action-state transitions observed during training. Since the environment was already discretized into cells, we didn't need to bin continuous values - instead, we counted transitions between specific grid cells, creating natural discrete distributions for our information calculations.

During the 200k-step training run, the robotic-arm agent exhibits clear, interpretable information-theoretic signatures that map onto the emergence of a coherent policy. Fig. 1 and Table 1 summarize the evolution of the core metrics. From Table 1 we can observe the following:

- State entropy H(S) grows from 3.39 bits at initialization to 4.44 bits at step 200,000 (+31 %). This reflects the agent's exploration of an increasingly diverse portion of the state space.

- Action entropy H(A) follows the opposite trend: it remains high early in training (≈8.4 bits) but declines to 6.68 bits by the end. The decrease signals that, as learning proceeds, the policy assigns higher probability mass to a smaller set of actions—i.e., the agent is committing to reproducible control commands.

- Mutual information between states and actions, MI(S;A), rises sharply from 0.84 bits to 2.83 bits (+238 %). Knowing the current state therefore becomes over three times more informative about the selected action, highlighting increased policy determinism even while state variety is expanding.

- Joint mutual information MI(S,A;S′) traces an inverted-U shape: it increases during the exploratory phase, peaks at 3.46 bits around step 80 k, and settles at 1.93 bits by step 200 k. This suggests that the agent first builds a broad (high-entropy) mapping from state–action pairs to next states, then prunes it down to a smaller subset of highly effective transitions as exploitation dominates.

- Finally, MI(A;S′) grows modestly—from 0.83 bits to 0.99 bits (+19%). Actions become somewhat more predictive of next states, but the effect is muted relative to MI(S;A); most of the predictive structure resides in the full state–action context, not in actions alone.

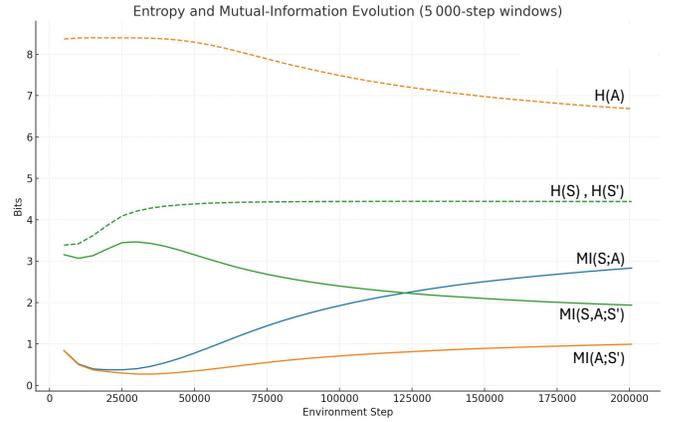

**Fig. 1.** Time–series of state, action, and next-state entropies (dashed curves) together with the three mutual-information terms (solid curves) for the 200 k-step learning run. Metrics are computed in 5,000-step cumulative windows, so each point reflects all data seen up to that step. The rise of MI(S;A) and MI(A;S′) signals growing policy coherence, while the gradual fall of H(A) shows the controller becoming more deterministic.

**Table 1. Quantitative summary of entropy and MI trajectories**

| Metric (bits) | Initial | Peak | Final | Δ (Peak – Initial) | Δ / Initial |
|---|---|---|---|---|---|
| H(S) | 3.385 | 4.444 | 4.439 | 1.059 | 31.30% |
| H(A) | 8.366 | 8.396 | 6.681 | 0.03 | 0.40% |
| H(S′) | 3.385 | 4.444 | 4.439 | 1.059 | 31.30% |
| MI(S;A) | 0.836 | 2.83 | 2.83 | 1.994 | 238.40% |
| MI(A;S′) | 0.829 | 0.991 | 0.991 | 0.161 | 19.40% |
| MI(S,A;S′) | 3.153 | 3.464 | 1.934 | 0.31 | 9.80% |

**Table 1**. Initial, peak, final values and relative changes (Δ / Initial) for all six metrics plotted in Figure X. Peaks correspond to maximum uncertainty (for entropies) or maximum predictability (for MI). Values are in bits; window size = 5000 steps.

### B. Correlation Between Information Metrics and Performance

The performance metrics in Figure 2 reveal a compelling correlation with the information patterns identified in Table 1. Average episode length (orange dashed line) and reward (blue solid line) show dramatic improvements during training, with distinct phases that align with our information-theoretic analysis.

During the early phase (0–25,000 steps), we observe the most dramatic performance changes. Episode length decreases rapidly from ≈ 450 steps to ≈ 10 steps, while average reward improves from –0.22 to ≈ –0.14. This period coincides with MI(S,A;S′) peaking at 3.464 bits, a phase in which state–action pairs are maximally informative about subsequent states, and the agent exhibits its most rapid performance gains.

The middle phase (25,000–75,000 steps) shows continued—but more gradual—improvement. Reward increases from –0.14 to ≈ –0.08, and episode length stabilizes around five steps. Over the same interval MI(S;A) continues to rise while MI(S,A;S′) declines, indicating a transition from broad exploration to policy refinement as the agent selects a narrower set of high-value state–action pairs.

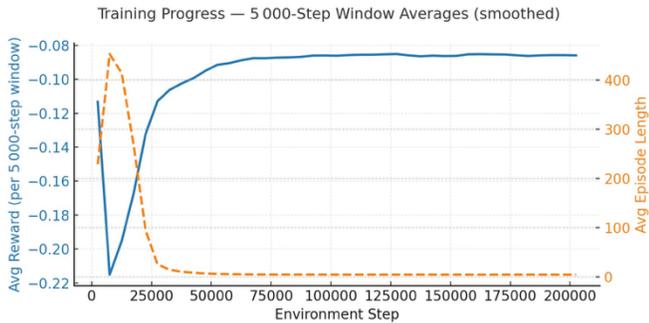

**Fig. 2.** Evolution of task performance during training. Average episode reward (solid blue) and average episode length (dashed orange) are computed in 5 k-step sliding windows across the 200 k-step learning run. Rewards become progressively less negative while episode length collapses from ≈ 450 to ≈ 5 steps, then both curves plateau after ~75 k steps—marking policy convergence.

The final phase (75,000–200,000 steps) displays performance stability: reward plateaus near –0.08 and episode length remains consistently low. Information metrics likewise converge—MI(S;A) reaches its maximum of 2.83 bits, and MI(S,A;S′) levels off at 1.934 bits—signaling that the learned policy has achieved a steady equilibrium between predictability and environmental dynamics.

The temporal alignment of information and behavioral curves suggests a coherent narrative of skill acquisition:

- Model construction—rising MI(S,A;S′) captures the agent's effort to learn dynamics.
- Policy sharpening—increasing MI(S;A) reflects decisive action selection.
- Performance consolidation—stable entropies and mutual information's parallel stable rewards.

### C. Learning Phases Revealed Through Information Dynamics

The information traces divide training into three readily identifiable stages that line up with the agent's behavioral progress.

1. Exploration (0–25,000 steps). State entropy H(S) climbs from 3.39 to roughly 4.2 bits as the agent discovers new parts of the state space, while action entropy H(A) remains high (≈ 8.3 bits) because many actions are being tried. During the same interval mutual information between the combined state-action pair and the next state, MI(S,A;S′), surges to its peak value of 3.46 bits, indicating that the agent is busily modeling how its moves affect future states. This burst of information coherence coincides with the steepest behavioral gains: average episode length collapses from about 450 to 10 steps and reward improves from –0.22 to –0.14.

2. Refinement (25,000–75,000 steps). Once the environment is broadly mapped, H(S) stabilizes near 4.4 bits, but H(A) starts to fall (≈ 8.3 → 7.5 bits) as the policy becomes less exploratory. Mutual information between states and chosen actions, MI(S; A), rises sharply (1.5 → 2.5 bits), meaning the agent's action selection grows tightly coupled to its current state. Meanwhile MI(S,A;S′) begins a gradual decline from its earlier peak, showing the agent is now following a narrower band of state transitions. Performance keeps improving, though more slowly:

rewards climb toward –0.08 and episode length settles near 5 steps.

3. Exploitation (75,000 – 200,000 steps). In the final stretch, H(A) drops further to 6.68 bits as the policy becomes decisively deterministic. MI(S;A) tops out at 2.83 bits, reflecting a strong one-to-one mapping from states to actions, while MI(S,A;S′) levels off around 1.93 bits because the agent repeatedly traverses a small set of efficient trajectories. Rewards and episode length plateau (≈ –0.08 reward, 3–4 steps), signaling policy convergence.

Taken together, these three phases illustrate how information-theoretic measures expose the hidden structure of learning: an initial period of broad hypothesis-building, followed by selective refinement, and ending with stable exploitation. The shift from high H(A) and rising MI(S,A;S′) to lower H(A) and maximal MI(S;A) provides a quantitative fingerprint of the classic exploration-to-exploitation transition.

The progression from exploration to exploitation reveals an important trade-off in learned policies. While the agent's increasing specialization—evidenced by rising MI(S;A) and declining MI(S,A;S')—enables efficient task performance, it also raises questions about robustness. The narrow, highly-optimized information channels that characterize the exploitation phase may be particularly vulnerable to disruption. To investigate this hypothesis and demonstrate the diagnostic power of information metrics, we conducted controlled perturbation experiments on the fully-trained agent.

### D. Vulnerability Analysis Through Information Signatures

Our learning phase analysis suggests that agents in the exploitation phase, with their specialized information pathways, might respond differently to various types of system degradation. Specifically, we hypothesize that: (1) the high MI(S;A) characteristic of specialized policies makes them vulnerable to disruptions in the state-action mapping, and (2) the reduced MI(S,A;S′) indicates fewer alternative pathways for achieving goals when primary channels are disrupted. To test these hypotheses and demonstrate how information metrics can diagnose system failures, we designed controlled perturbation experiments targeting different information channels.

### E. Deployment Perturbation Analysis

Having established that the exploitation phase creates specialized but potentially fragile information channels, we tested how different types of perturbations disrupt these channels in the fully-trained agent

To evaluate the robustness and diagnostic power of our information-signature framework under real-world conditions, we ran two post-training deployment experiments in which a single, controlled noise perturbation was applied at step 10,000. In Run 1, we injected Gaussian noise into the policy's action outputs to mimic actuator degradation; in Run 2, 3 we injected noise into the state observations to mimic sensor faults. These experiments test two critical hypotheses: that our entropy and mutual-information metrics will sensitively detect the onset of distributional drift, and that the distinct information channels will localize the disturbance to either the action or observation pathway.

As shown in Figure 3, before the 10 k-step injection both runs show stable information signatures: all three entropies (dashed lines) plateau and the four mutual-information curves (solid lines) sit tightly stacked, indicating a coherent and fully learned control loop. Once action noise is introduced (upper chart) the entropies drift only slightly, but the mutual-information terms dip in unison and then level off—revealing that actuator jitter blurs the relationship between successive states without completely disrupting it. In contrast, observation noise (lower chart) triggers an abrupt, much deeper collapse of every MI curve while leaving the entropies comparatively intact; the network can no longer predict the next state from the corrupted input, even though its internal variability remains high. The resulting "fingerprints" are therefore distinctive: a modest, parallel MI drop flags actuator degradation, whereas a steep MI plunge that outpaces any entropy change signals sensor or perception faults. This qualitative separation lets the monitoring system localize both the source and relative severity of a disturbance without relying on raw reward traces.

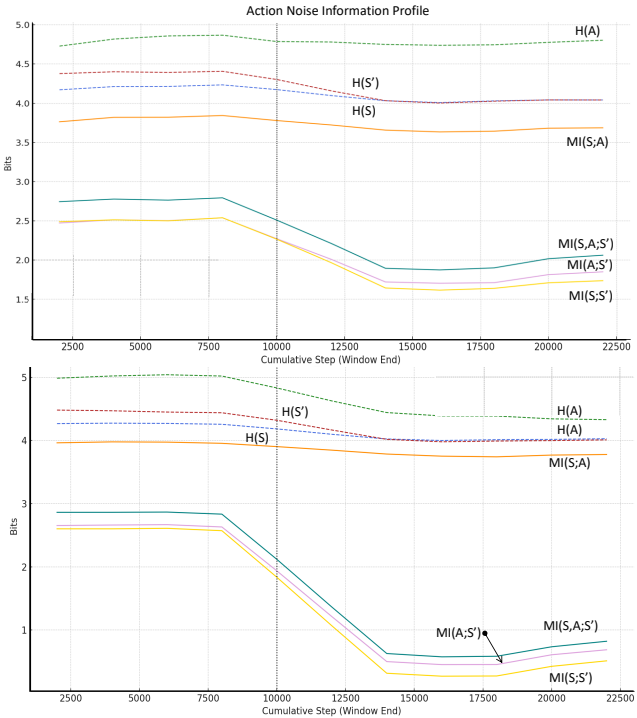

**Fig. 3** *Upper:* action-noise run; *lower:* observation-noise run. Dashed lines are entropies H(S), H(A), H(S′); solid lines are mutual-information measures MI(S;A), MI(A;S′), MI(S,A;S′), MI(S;S′). Curves are smoothed 2 k-step windows; the dotted vertical line marks the 10 k-step onset of noise. Action noise causes a modest, parallel drop in all MI terms while entropies drift slightly, whereas sensor, or states, noise produces a sharper MI collapse (bellow 1 bit) with moderate change in entropies—yielding distinctive profiles that localize both the source and severity of the disturbance.

This pattern extends to next-state predictive information, how much knowing the current state predicts the next state, regardless of the actions, MI(S;S′) and support-size dynamics (Fig. 4). Action noise reduced predictive information by 0.80 bits, whereas observation noise erased 1.83 bits of "memory" (Fig 3; Table 2). Unique-action counts collapsed from ~150 to

~61 in both runs, but unique-state counts held at ~27 bins under action noise while transiently rising to ~38 bins under observation noise (Table 9). Thus, corrupting sensor inputs not only compresses action distributions but also scatters the agent through spurious state regions, amplifying collapse in both entropy and mutual-information channels. In addition, Figure 8 reveals that MI(S; S′) does not merely signal the *presence* of actuator faults; its nadir depth (Δ0.80 bits for σ² 0.1 vs. Δ1.32 bits for σ² 1.0) scales monotonically with noise variance. Hence the metric provides a quantitative estimate of fault *severity*, complementing the source-specific signatures.

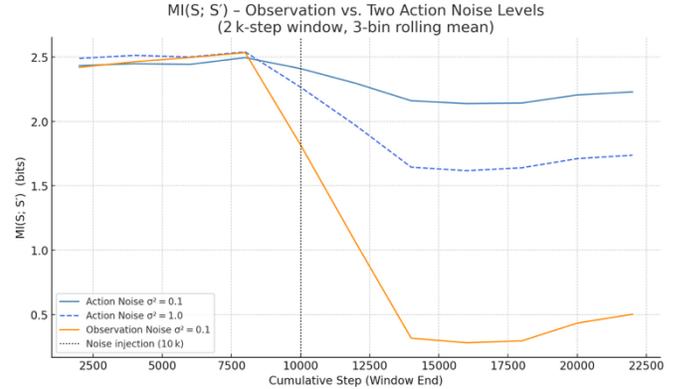

**Fig. 4** MI(S; S′) sensitivity to action-space noise magnitude. Solid lines = σ² 0.1 (blue) and σ² 1.0 (orange); dashed = observation noise σ² 0.1 for reference. Curves computed with a 2 k-step sliding window, 3-bin rolling mean. Higher noise produces proportionally deeper and longer MI(S; S′) collapses, demonstrating that the metric encodes both fault type and degree.

Together, these results demonstrate that monitoring multiple, disaggregated information pathways allows not only detection of distributional drift but—crucially—localization of its source. A minor drop in downstream predictability with intact state–action coupling flags actuator faults, whereas simultaneous, deep collapses across all metrics expose sensor corruption. To our knowledge, no existing drift-detection method offers this level of diagnostic precision, making the information-signature framework a novel tool for resilient, self-correcting autonomy.

## V. DISCUSSION

### A. Efficiency and Specialization in Successful Policies

Our results show that the agent's policy does not simply maximize information; instead, it becomes increasingly efficient—retaining only the bits that matter for high-performance control.

- Growing selectivity in state → action mapping. State entropy H(S) plateaus at ≈ 4.44 bits, yet MI(S;A) keeps rising, ending at 2.83 bits (≈ 64 % of H(S)). By late training, the agent can infer almost two-thirds of its action choice directly from the current state—evidence of a near-deterministic, task-focused policy.

- Pruning of state–action–next-state trajectories. MI(S, A; S′) climbs quickly, peaking at 3.46 bits around 20 k steps as the learner builds a broad model of environment dynamics. Thereafter it drops to 1.93 bits, showing that many exploratory transitions are abandoned in favor of a small set of reliable, high-value paths.

- Behavioral payoff. As informational overhead shrinks, episode length falls from ~450 to ≈ 3.9 steps, and average reward stabilizes near −0.08. The agent now reaches the goal through the shortest, most predictable sequence of states.

These twin trends—higher MI(S;A) coupled with lower, stabilized MI(S,A;S′)—mirror human skill acquisition: novices survey a wide solution space, whereas experts act through streamlined, highly informative cues. Optimal policies therefore emphasize information efficiency, not sheer information volume: maximize the relevance of each bit (tight state-action coupling) while minimizing unnecessary processing (discarding low-value transitions). This balance offers a richer criterion for judging policy quality than performance scores alone.

Our perturbation experiments show that the same selective information processing responsible for policy efficiency also governs the agent's response to external noise. Under perturbation conditions, the agent's action and state distributions rapidly compress, reflecting a fallback into fewer, more reliable behaviors. For example, injecting noise directly into the agent's observations reduced state entropy and action entropy, highlighting the agent's immediate narrowing of behavioral scope to maintain reliable operation. In contrast, noise added to actions alone caused more modest entropy reductions. This response demonstrates that successful policies exhibit a built-in mechanism to revert to a more specialized, reliable subset of their full action repertoire when faced with unexpected conditions, thus preserving operational coherence under stress.

### B. Information Signatures of Deployment Drift

To assess how effectively information-theoretic metrics diagnose deployment-time drift, we analyzed policy responses under controlled perturbations introduced at step 10,000. Our results demonstrate clear, distinguishable signatures depending on whether noise was injected into the action outputs or the state observations.

When action-space noise was introduced (Run 1), mutual information between states and actions (MI(S;A)) remained stable or even slightly increased (+0.003 bits), suggesting the policy maintained or tightened its state-to-action mappings. However, predictability from actions to next states, MI(A;S′), decreased by 0.177 bits, and joint mutual information MI(S,A;S′) decreased by 0.119 bits, reflecting reduced reliability in action outcomes. State entropy (H(S)) and action entropy (H(A)) experienced modest decreases (−0.034 and −0.049 bits respectively), signaling that the agent slightly narrowed its behavioral repertoire without significantly changing its state-space footprint.

In contrast, observation-space noise (Run 2) induced far more severe disruptions across all metrics. State entropy H(S) dropped by 0.093 bits, action entropy H(A) by 0.218 bits, and next-state entropy H(S′) by 0.170 bits, reflecting substantial compression in state and action distributions. Mutual-information metrics fell dramatically: MI(S;A) decreased by

0.058 bits, MI(A;S′) collapsed by 0.668 bits, and MI(S,A;S′) fell by 0.589 bits. Predictive information MI(S;S′), a measure of temporal coherence, saw a particularly pronounced decline (−1.83 bits), indicating severely impaired short-term predictability.

These results highlight that our information-theoretic metrics do more than detect generic shifts: they uniquely pinpoint whether disturbances originate from sensor-level observations or actuator-level outputs. Unlike standard drift-detection methods, which signal only that something has changed, our approach localizes the source of drift, thus enabling more targeted diagnostics and interventions.

### C. Comparison to Existing Drift-Detection Methods

Standard drift-detection methods, such as CUSUM [28], ADWIN [6], and KL-divergence or Maximum Mean Discrepancy (MMD) tests [15], effectively signal distributional shifts by tracking changes in statistical properties of observations or outputs. However, these methods typically provide no mechanism to pinpoint the underlying source of the shift—whether it originates in sensor inputs or actuator outputs.

Our information-theoretic framework offers a significant advance in this respect. By separately tracking state-action, action-next-state, and state-next-state information pathways, we gain the unique ability to diagnose precisely whether perturbations impact input perceptions or output actions. This diagnostic precision exceeds traditional drift-detection capabilities, providing an essential tool for targeted and effective intervention in closed-loop reinforcement-learning systems.

### D. Selective Information Encoding as a Signature of Effective Policies

Our results reveal a fundamental pattern in how reinforcement learning agents develop representations: mutual information between states and actions increases even as the agent encounters more diverse states. This finding challenges the intuitive assumption that learning merely reduces uncertainty. Instead, effective learning involves developing increasingly selective attention to task-relevant patterns within growing environmental complexity.

This selective information encoding manifests as the agent's ability to extract meaningful structure from increasingly varied observations, strengthening state-action relationships despite higher state entropy. The observed increase in MI(S;A) from 0.836 to 2.83 bits represents a dramatic 238.4% growth, while the ratio of MI(S;A) to state entropy increases from 24.7% to 63.8%. This indicates that the agent progressively channels more environmental information into its action selections, ignoring irrelevant variations. This pattern aligns with Tishby and Zaslavsky's [37], information bottleneck perspective, where neural networks optimize the tradeoff between compression and prediction, but extends this view to reinforcement learning contexts.

Particularly revealing is the inverted U-curve we observed in MI(S,A;S′), which peaks during early learning before gradually declining. This pattern suggests a progression from broad exploration of state-action-state relationships to selective optimization of the most effective transitions. As the agent develops expertise, it narrows its focus to a more efficient subset of state-action-state mappings, similar to how human experts develop streamlined, efficient approaches to familiar tasks rather than continuously exploring all possibilities.

The development of selective attention provides a theoretical explanation for why policies often fail when deployed in subtly different environments: if the agent has learned to selectively encode specific patterns that no longer apply in the new environment, its information processing pathway breaks down. By monitoring MI(S;A), we can potentially identify when an agent's selective information encoding no longer matches environmental reality, providing an early warning system for policy degradation before performance metrics decline. This perspective offers a more fundamental understanding of representation quality than reward-based metrics alone [25].

### E. Information Metrics as Architecture-Agnostic Quality Indicators

A significant advantage of information-theoretic metrics is their applicability across diverse architectures and algorithm types. Unlike specialized representation learning approaches that require specific network designs or auxiliary objectives, mutual information calculations depend only on observable state-action distributions, making them compatible with any policy implementation. This universality enables comparative analysis across different algorithms and architectures using a common framework.

Information metrics also offer interpretable units (bits) with consistent meaning regardless of application domain. This contrasts with performance metrics like reward, which are task-specific and often incomparable across environments. As Zhang et al. [41], note, performance metrics can mask underlying problems with representation quality until failure occurs, while information measures potentially provide earlier indicators of misalignment. The consistency of information as a metric enables principled threshold setting for detecting distribution shifts across diverse deployment settings.

Furthermore, these metrics align with fundamental information processing constraints that apply to both artificial and biological learning systems [13]. The observed patterns in mutual information capture how effectively agents extract and utilize environmental structure—a fundamental aspect of intelligence that transcends specific architectural implementations. This perspective offers a unifying framework for analyzing how learning systems develop effective representations across sensory, decision-making, and motor systems, potentially bridging reinforcement learning with broader cognitive science insights.

### F. Limitations, Assumptions, and Design Trade-offs

Our analysis focuses on detailed characterization of a single representative run. While we verified consistent patterns across 4 independent runs, formal statistical analysis across runs remains for future work. Despite this limitation, the clear and substantial changes in information metrics (e.g., 238% increase in MI(S;A)) provide strong initial evidence for our framework's utility.

While our information-theoretic approach offers valuable insights, several limitations and assumptions deserve consideration. First, calculating mutual information requires reliable state and action distribution estimates, which become challenging in high-dimensional continuous spaces. Our discretization approach introduces estimation noise that varies with binning resolution—finer binning better captures subtle relationships but requires more data, while coarser binning provides more stable estimates but may miss important patterns. This trade-off represents a fundamental challenge in applying information theory to complex systems [7].

Second, our framework assumes a degree of stationarity in the underlying task. While MI(S;A) can detect distribution shifts, significant non-stationarity in the environment may require adaptive baseline updates to distinguish problematic shifts from expected variations. As Yang et al. [40] note, distinguishing harmful distribution shifts from benign ones remains challenging for any detection method. Incorporating task-specific knowledge about acceptable variation ranges could enhance the practical utility of information metrics in deployment scenarios.

Third, our approach currently analyzes single-agent systems with fully observable states. Extending to multi-agent settings would require accounting for information flow between agents, potentially incorporating concepts from multi-agent information theory [4]. Similarly, partial observability introduces additional challenges in estimating true information relationships, as observed patterns may reflect incomplete information rather than actual policy quality. Combining our approach with hidden state estimation techniques could address these limitations in future work.

Finally, while we demonstrate information metrics' value for detecting misalignment, developing appropriate adaptation strategies based on these signals remains an open challenge. Information gradients offer promising directions for targeted parameter updates, but connecting detection to effective adaptation requires further investigation.

### VI. CONCLUSION AND FUTURE WORK

This paper introduces a novel information-theoretic approach to analyzing representation quality in reinforcement learning agents. By tracking mutual information between states and actions, we reveal a fundamental pattern in effective learning: agents develop increasingly selective attention to task-relevant patterns amid growing environmental complexity. This selective information encoding manifests as increasing MI(S;A) despite rising state entropy, providing a quantitative signature of representation quality that complements traditional performance metrics.

The diagnostic utility of information metrics extends beyond analyzing learning dynamics to detecting potential policy degradation before performance visibly declines. By establishing baseline information patterns during successful operation, deviations from these patterns can serve as early warning signals when an agent's representation no longer aligns with its environment. This capability addresses a critical gap in current RL systems, which typically lack intrinsic mechanisms for detecting their own misalignment with changing conditions.

Future work could explore integrating these information metrics directly into RL algorithms to create self-adaptive systems capable of detecting and responding to distribution shifts without human intervention. By using information gradients to guide targeted parameter updates, agents could potentially maintain alignment with changing environments without requiring complete retraining. This approach would help close the supervision gap in deployed systems by enabling continuous self-assessment and adaptation.

Additionally, extending this framework to partially observable and multi-agent settings represent a promising direction for future research, potentially revealing new insights into coordination and representation sharing across agent networks. Information-theoretic measures may also offer new perspectives on transfer learning by quantifying how effectively representations maintain their information relationships across different tasks and environments.

By grounding representation quality in fundamental information-theoretic principles, this work takes a step toward more robust, self-monitoring RL systems capable of maintaining reliable performance in dynamic, real-world environments.

## ACKNOWLEDGMENT

The authors used Claude (Anthropic) and ChatGPT (OpenAI) as linguistic and stylistic tools to enhance the clarity, grammar, and scientific tone of the manuscript. All content and scientific statements remain the sole responsibility of the authors.